\documentclass[conference]{IEEEtran}
\IEEEoverridecommandlockouts
\usepackage{cite}
\usepackage{amsmath,amssymb,amsfonts,amsthm}
\usepackage{algorithmic}
\usepackage{graphicx}
\usepackage{textcomp}
\usepackage{xcolor}
\usepackage{hyperref}
\usepackage{float}

\newcommand{\EMPH}[1]{\textbf{\underline{#1}}}

\def\BibTeX{{\rm B\kern-.05em{\sc i\kern-.025em b}\kern-.08em
    T\kern-.1667em\lower.7ex\hbox{E}\kern-.125emX}}

\begin{document}

\title{Tensions Between the Proxies of Human Values in AI}


\author{\IEEEauthorblockN{Teresa Datta*, Daniel Nissani*, Max Cembalest, Akash Khanna, Haley Massa, John Dickerson}
\IEEEauthorblockA{
Emails: \{teresa, daniel.nissani, max.cembalest, akash, haley, john\} @arthur.ai}
\IEEEauthorblockA{Arthur}}


\maketitle

\def\thefootnote{*}\footnotetext{Equal Contribution}

\begin{abstract}
Motivated by mitigating potentially harmful impacts of technologies, the AI community has formulated and accepted mathematical definitions for certain pillars of accountability: e.g. privacy, fairness, and model transparency. Yet, we argue this is fundamentally misguided because these definitions are imperfect, siloed constructions of the human values they hope to proxy, while giving the guise that those values are sufficiently embedded in our technologies. Under popularized methods, tensions arise when practitioners attempt to achieve each pillar of fairness, privacy, and transparency in isolation or simultaneously. In this position paper, we push for redirection. We argue that the AI community needs to consider all the consequences of choosing certain formulations of these pillars---not just the technical incompatibilities, but also the effects within the context of deployment. We point towards sociotechnical research for frameworks for the latter, but push for broader efforts into implementing these in practice. 

\end{abstract}


\section{Introduction}
\vspace*{10px}
High profile events continue to spur popular discourse on the definition of, the need for, and the limitations placed on ``responsible AI.'' Ranging from Latanya Sweeney's re-identification of individuals with public datasets in 1997 \cite{reidentification} to ProPublica's finding that a popular recidivism risk scoring algorithm was heavily biased towards Black people in 2016 \cite{angwin_larson_kirchner_2016}, the public has grown increasingly aware that AI systems need to be held to account \cite{fairml}.

\begin{figure*}
    \centering
    \includegraphics[scale=0.23]{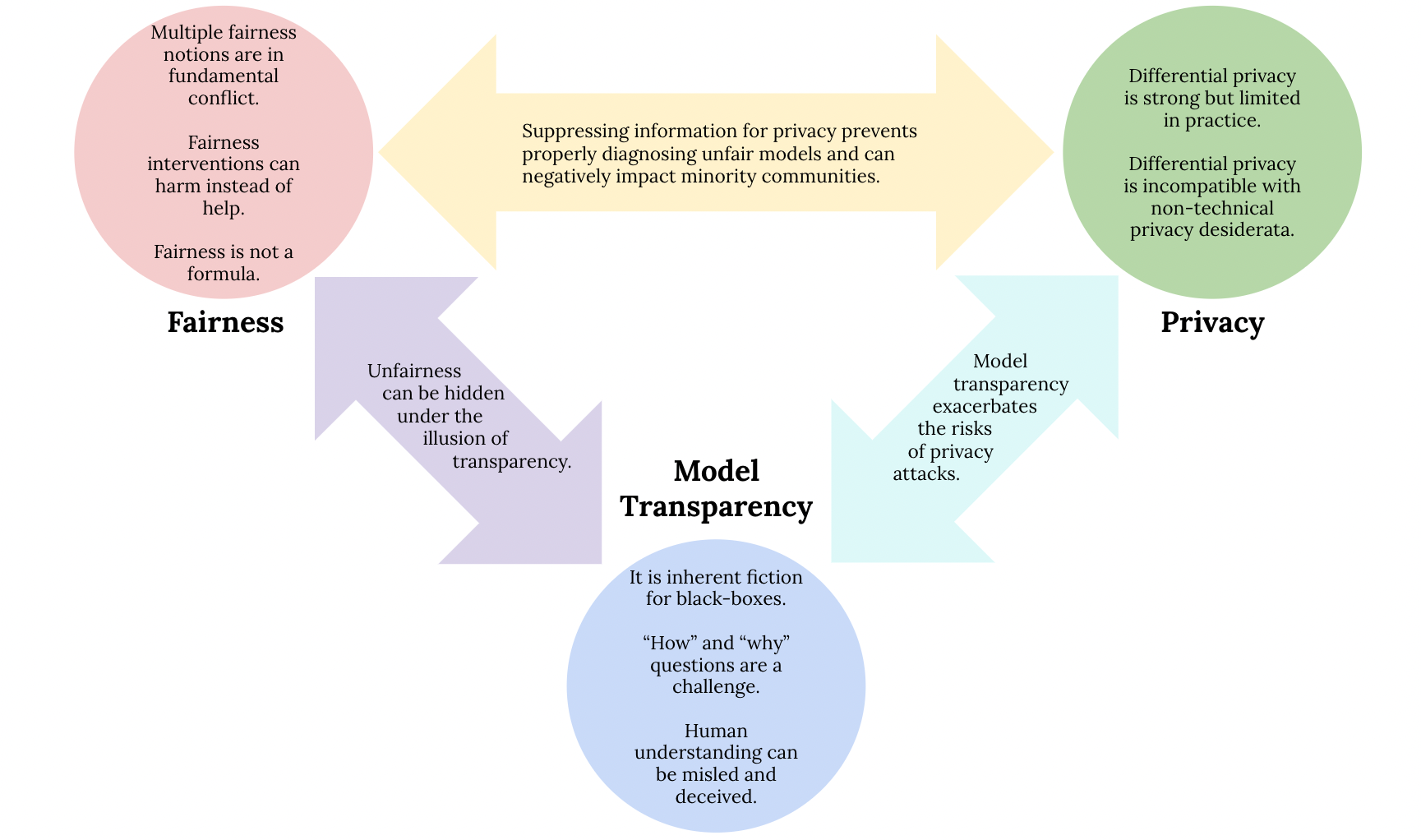}
    
    \textit{Figure 1: A summary of the tensions we identify within and between popular proxies of human values. Incorporating any one of these pillars is itself a challenge, and incorporating them simultaneously requires handling competing priorities.}
    \label{fig:triangle}
\end{figure*}
In an effort to incorporate our human values related to privacy, fairness, and model transparency, the AI community has adopted automatable, domain-agnostic mathematical formulations. 
Consider \EMPH{fairness}: over the past decade, the fairness in machine learning community has come up with various definitions to combat unfavorable imbalances in model predictions towards minoritized groups \cite{fairml}. Take \EMPH{privacy}: since 2006, the privacy community has heavily leaned on differential privacy \cite{diff_priv_sem}, a probabilistic guarantee that a model or summary statistic won't change based on the perturbation of a single data point. And, perhaps most contentiously within the AI/ML community itself, consider \EMPH{model transparency}: some in the community have allowed that deployed models' behavior can be explained in a black-box, model agnostic way, via interpretable surrogate models such as LIME or SHAP \cite{lime,shap}. Others in the community argue strongly against this approach as critically flawed, instead proposing the top-level intervention that \emph{only} inherently interpretable models be deployed \cite{blackboxbad,mythos}.  Indeed, this particular dimension continues to struggle with defining what makes an explanation ``good,'' or even what makes an explanation an explanation \cite{defineinterp,molnar2022}. This paper recognizes that ``explainability" is an overloaded term. Thus, ``model transparency" will refer to the techniques for understanding model decisions, specifically inherent interpretability and post-hoc explainability techniques. We describe these proxies as mathematical and technical because measuring their success is often framed via metrics that can be explicitly calculated and optimized.  

These technical proxies of core value pillars are not only relevant from a moral or technical standpoint, but also from a regulatory perspective. We focus the scope of this work on tensions between human values and their technical proxies, although we acknowledge that much work needs to be done to align the research community with the practical considerations of the goals of regulators and policy makers \cite{kumar2022equalizing}.

\vspace*{10px}

\textit{We must examine the consequences of our formalizations.}

\vspace*{10px}

We must acknowledge that implementing these specific formulations into technologies is a \textbf{choice}, and any choice will have consequences. We outline three categories of tensions that arise:
\begin{enumerate}
    \item Tensions within the value pillar.
    \item Tensions with other value pillars.
    \item Tensions with the real world context of deployment.
\end{enumerate}

The first is the inherent inconsistencies within the value pillar that these formalizations warrant. As an example, current fairness definitions are unable to be simultaneously enforced in a machine learning model \cite{kleinberg} and force practitioners to choose one. The second source of tension arises from the compounded impossibility of fully operationalizing another value pillar, such as explainability techniques hindering the privacy of algorithms \cite{shokri2021privacy}, while natively interpretable methods may have adverse impacts on marginalized groups \cite{Meng22:Interpretability}. Figure \ref{fig:triangle} outlines the technical tensions identified within and between fairness, privacy, and model transparency. 

Most importantly, there are the consequences that arise in the context of deployment. What are the effects of implementing these value choices in real-world sociotechnical systems involving a complex interplay of technical and human actors? We recommend frameworks from the Science, Technology, and Society (STS) field, such as substantive algorithmic fairness\cite{escaping_impossibility}, contextual integrity\cite{privacy_in_context}, and domain-specific transparency methods, to address this vital area of consideration. However, we also acknowledge that while there are many resources to address the first two categories of consequences, there are not enough exemplars of how technologists can consider the ramifications of the choices they make (which they often do in siloed, context-agnostic settings). We push for further inquiries into addressing this last gap. We should not be examining accountability metrics in theoretical silos, but rather within specific domains.

\vspace*{10px}

\textit{Paper layout}

\vspace*{10px}

In Section \ref{sec:pillars}, we discuss the inconsistencies within popular technical proxies of fairness, privacy, and transparency and their human values. In Section \ref{sec:tensions}, we survey these pillars pairwise and discuss how the intersections of these proxies result in even greater tensions. In Section \ref{sec:forward}, we examine the final tension and the implications of understanding the sociotechnical system a technology may be deployed into. We motivate the requirement for context-based formulations of our three pillars, and push for greater research and industry focus into these areas.


\section{Tensions within pillars}
\vspace*{10px}
\label{sec:pillars}


\subsection{Fairness} 

\vspace*{10px}

\textit{Popular formulations target different notions of fairness and do not work together.}

\vspace*{10px}

In response to concerns over potential discriminatory impacts of algorithmic decision making, over 21 technical formulations of fairness have been defined \cite{Verma2018,21fair}. Table \ref{tab:fair_defs} in Appendix A outlines five popular metrics. These formulations aim for different notions of fairness, largely categorized as independence, separation, and sufficiency. Unfortunately, these criteria cannot all be achieved in a single model without either having a perfect or trivial classifier \cite{fairml, kleinberg}. This finding is fundamentally concerning. Each of these notions of fairness embeds a specific way in which the human value of fairness is conceptualized. And if it is impossible to achieve multiple forms of fairness at the same time, then can any system ever be considered fully fair? And if this is the case, then why are these still the metrics that we use for evaluation.

\vspace*{10px}

\textit{Fairness interventions can cause harm.}

\vspace*{10px}

These criteria are not implemented in vacuums, but rather in dynamic, real world systems. When these fairness interventions are put in place, they may be ``overeager" and propagate long-term harms to the underserved groups they hope to benefit due to a lack of consideration of long-term well-being \cite{delayed_impact}. Moreover, \cite{nilforoshan} shows that causal notions of fairness, including equalized odds, are Pareto dominated in a system, meaning that for each fairness definition there exists a better classifier that achieves better accuracy and better outcomes for protected groups.

\vspace*{10px}

\textit{There are alarming mismatches with nontechnical conceptions of fairness.}

\vspace*{10px}

These fairness definitions do not always resonate with how people outside of the AI community---most of the population---think about practical fairness. We must recognize that the tools we build affect everyone, and this necessitates a democratic duty to consider value pillars with public comprehension and sentiment in mind. Public understanding and acceptance of formalizations of ``what is ethical?'' should be more highly prioritized. When non-technologists are asked how they feel about these definitions, not only do they have trouble fully comprehending them \cite{nonexpert_comprehension}, but they also do not agree with all of them \cite{fairness_fares}. In fact, in some cases, greater \textit{comprehension} of the fairness metrics, specifically demographic parity, is actually associated with \textit{increased negative sentiment} about that metric \cite{nonexpert_comprehension}. 

Of course, AI researchers are not the first to be invested in conceptualizing what it means for something to be fair. Philosophers have a long history of grappling with fairness \cite{rawls1991justice}, and economists have been forced to examine the implications of equity in practice \cite{young1995equity}. When fair ML research is surveyed under the lens of political philosophy \cite{binns}, mismatches are noted in how the community conflates terms like ``discriminatory" and ``unfair", and how even the use of the term ``fairness" functions as a catchall for a diverse set of ``normative egalitarian considerations."

\subsection{Privacy} 
\theoremstyle{definition}
\newtheorem{definition}{Definition}[section]

Born out of a necessity to reveal statistics about a population without allowing access to information about individuals \cite{diff_priv_sem}, differential privacy has become the canonical notion of privacy in the AI community. Once achieved for some $\epsilon$, differential privacy provides a probabilistic guarantee that a machine learning model will perform the same if a single data point is removed or replaced. Appendix \ref{sec: defprivacy} outlines these formal technical definitions.








Although differential privacy offers a rigorous guarantee on an individual data point's privacy, it does have its limitations. Even for a model that achieves differential privacy, the more the model overfits, the more susceptible it is to membership and attribute inference attacks \cite{priv_of}. Also, data points that are not represented well in a dataset, such as outliers, either incur large privacy costs (i.e. large $\epsilon$) \cite{priv_out} or are memorized by the model and can be exposed much more easily given certain prompts \cite{carlini_2019}.

\vspace*{10px}

\textit{There are conflicts with nontechnical notions of privacy.}

\vspace*{10px}

Differential privacy is a very specific notion of privacy that the AI community has adopted as a gold standard. However, it is not a complete account of privacy, and does not address issues of collection and usage of personal data. In fact, according to a Pew Research poll, 64\% of US adults are not too or not at all comfortable with their personal data being shared with outside research groups for the improvement of society \cite{PEW1}. The ML research community's conceptualizations of privacy do not consider how this term is used publicly. Furthermore, there is limited transparency and understanding of privacy techniques and policies. According to the same PEW research poll, a majority of Americans say they have little to no understanding of existing data protection techniques or laws \cite{PEW1}. 

\vspace*{10px}

\textit{Differential privacy is not suited for non-tabular data.}

\vspace*{10px}

Much of the limitations of differential privacy that we have described thus far pertain to tabular data. But what about other data types? Let's take unstructured text as an example. Although there has been some success in building differentially private large language models (LLMs) \cite{diff_priv_bert}, the very notion of what privacy means for an LLM is ill-defined. An initial definition for LLM privacy is ``[t]o claim a language model is privacy preserving, it must only reveal private information (aka ``secrets'') in the right contexts and to the right people'' \cite{brown2022does}. The idea of privacy for language requires knowing who is receiving information, who is giving information, the context around why information has to be secret, and how a secret relates to all individuals involved either directly or indirectly. Thus, if we hope to extend privacy to other data domains, such as language, we need to have more robust and contextual definitions of privacy.


\subsection{Model Transparency}

Transparency refers to approaches or techniques devised to build trust and understanding in a model’s decisions \cite{weller_transparency}. Using a simple and interpretable model, when possible, offers a layer of transparency \textit{beyond} datasheets and model cards \cite{datasheets, modelcards}, because having an understanding of a model via interpretable model weights makes detecting and mitigating issues of performance and fairness easier. 

\vspace*{10px}

\textit{Transparency of black boxes is an inherent fiction using rough local approximation.}

\vspace*{10px}

Using a black-box model and explaining it post-hoc has become an increasingly popular approach in ``explainable AI" (XAI) because all that is needed is the model's inputs and outputs \cite{lime, shap}. However, there is significant debate as to how ``transparent" it really is to explain a model this way. All of these methods ``explain" by fitting a local surrogate model, drawing conclusions about the black-box model from the surrogate. These are all merely different types of  \textit{local-function approximation}. It has been proven that the locality of these explanations constrains them from being able to generate optimal global explanations \cite{lfa_lunch}. There is no way to have a ``ground truth" explanation if there can be no guarantees of these explanations representing the black-box completely and with fidelity\cite{blackboxbad}.


But why are black-box models and post-hoc explainability techniques prevalently used? First there is a notion of a tradeoff betweeen interpretability and model performance (although, the existence of such a tradeoff is questioned \cite{acc_exp_tradeoff_doesnt_exist}). Second, many practitioners perceive hard-to-interpret models as easier to use off-the-shelf, even though interpretable alternatives exist \cite{simple_models_exist}. This normalizes stakeholders to non-interpretable models \cite{blackboxbad}. However, ``inherently interpretable" models can have their own challenges. They can have an over-abundance of features or over-engineered features that make them hard to directly interpret \cite{mythos}. But they should be considered more often due to the accountability they provide, and black-box models should be treated with more skepticism despite the supposed post-hoc ``transparency." 

\vspace*{10px}

\textit{We don't know what is actually needed from transparency.}

\vspace*{10px}

There are a range of downstream tasks that transparency should enable for model designers, such as debugging, ensuring compliance with regulations, and generating hypotheses \cite{interp_stakeholders, explainable_ai_theory_driven}. Human-centered evaluations focus on user studies to examine whether explanations are actually helpful for humans in practical real-world use cases \cite{chandrasekaran-etal-2018-explanations, NEURIPS2020_adebayo}. 

The meaning of ``transparency" has extended into communicating \textit{how} and \textit{why} a model transforms particular inputs into the resulting outputs \cite{defineinterp, weller_transparency, molnar2022}. One could explain a neural network by reporting all of its weights, or one could explain it by visualizing activations of hidden layers \cite{olah2018}. When these methods provide too much information, simpler, contrastive, and sparse outputs are needed for \textit{comprehensibility} \cite{molnar2022}. Overall, there are insufficient formalisms at the moment for measuring the quality of explanations for practical use in machine learning.


\vspace*{10px}

\textit{Transparency can mislead.}

\vspace*{10px}

Explanation methods can mislead their intended audience, even when they are properly trained, by providing explanations that align with user's opinions. This leads to misplaced trust in faulty models due to confirmation bias---data scientists may overtrust and misuse interpretability tools without an accurate understanding of their output \cite{interpinterp}. In addition, transparency can mislead when the explanation algorithm distracts from what the model is directly doing. For example, using attention mechanisms may highlight associations entirely unrelated to a model's output \cite{deceive_attention}. Finally, some claim this whole pursuit of post-hoc explainability is completely misguided \cite{blackboxbad}, since explanations can never be completely faithful to the black box without being equal to the black box. They liken the practice of dissecting the meanings of explanations we don't understand to reading tea leaves \cite{tea_leaves}.

\section{Tensions between pillars}
\vspace*{10px}
\label{sec:tensions}
\subsection{Fairness and Privacy} 


\vspace*{10px}

\textit{Promoting privacy can harm fairness and vice versa.}

\vspace*{10px}

Differential privacy practices can amplify model unfairness \cite{priv_disp_imp} by reducing the accuracy disproportionately for underrepresented classes. Likewise, when models are fairness constrained, the data of minority groups in the training set can have a disproportionate impact on the model’s behavior and are thus often more susceptible to information leakage \cite{priv_risk_fair}.

There are also theoretical incompatibilities between fairness and privacy. In \cite{Cummings2019}, the authors show that under the constraints of differential privacy, exact statistical notions of fairness (Equality of False Positives and Equality of False Negatives) are unattainable. In \cite{fair_priv_tradeoffs} an impossibility theorem is introduced, proving that attempts to create a binary classifier that satisfies $\epsilon$-differential privacy and popular notions of fairness (Demographic Parity, Equalized Odds, and Equal Opportunity) could only result in a trivial classifier. 

\vspace*{10px}

\textit{Implementing differential privacy can negatively impact minority communities.}

\vspace*{10px}

Implementing differential privacy techniques has been shown to disproportionately impact minority communities.
This exact scenario arose with the inaugural employment of differential privacy for the 2020 US Census \cite{census_overview}. Published Census data has real-world consequences in the apportionment of over hundreds of billions of dollars in federal funding \cite{Bureau}, our understanding of health disparities \cite{Santos-Lozada_Howard_Verdery_2020}, and national confidence in governmental procedures due to historic undersampling of minority communities \cite{long_history}. The implementation of differential privacy was found to decrease the population of Native American reservations with fewer than 5000 people by an average of 34 percent \cite{akee2020importance}. The error between actual and differentially-privately-reported populations can result in dramatic differences in their allotment of federal funding, and could decide whether they are able to ascertain the funding for a road to a nearby town, or even a new school \cite{Wezerek_Riper_2020}. These smaller communities being subject to more erroneous representation has downstream allocation and representation implications, which is inherently an issue of equity and equal representation.

Legally and practically, notions of privacy
and fairness can be at odds. The Equal Credit Opportunity Act (ECOA) and associated Regulation B
control how a creditor can collect data on individuals. Namely, a creditor is not allowed to collect
demographic information related to a credit transaction \cite{kumar2022equalizing}. "Protecting" sensitive attributes by not
collecting them (similar to the idea of fairness through unawareness) actually supports discrimination
in the mortgage industry today. In \cite{martinez_kirchner_2021}, the authors outline how the fear of re-identification attacks has
banned the collection of credit scores, which results in ongoing racial discrimination as seen via the
public data mandates of the Home Mortgage Disclosure Act.

\subsection{Privacy and Explainability}
\vspace*{10px}

\textit{Privacy and Transparency have opposite goals.}

\vspace*{10px}

There are inherent tensions between an individual's right to privacy and transparency. In a responsible algorithmic system, a single user expects their data to be accessible by them, but to be secure or obfuscated to others. However, they also expect the ability to understand how their data was used to make decisions about them \cite{banisar2011right}. 

Explanations and interpretations inherently reveal information, and there are privacy tradeoffs when these are surfaced to external stakeholders. Comparisons can be drawn to privacy and transparency in clinical studies, where researchers want to present trustworthy results while protecting patient trial information \cite{medical_tensions}. Providing explanations for subsets of people, or even unique individuals, illuminates model behavior at the cost of exposing sensitive information. 


Ensuring trustworthiness in explanations can be difficult in systems that maintain privacy through data obfuscation. Masking sensitive attributes or adding noise to features inherently obscure data to human stakeholders, which could be seen as techniques used to manipulate results or change explanations. Recent research has supported the existence of a trade off between user privacy and model transparency \cite{dfp_explanations, dfp_interpretability_case, inter_dfp_pred}. 


\vspace*{10px}

\textit{Post-hoc explainers make models more susceptible to privacy attacks.}

\vspace*{10px}

Research has shown how providing model predictions along with feature based explanations leaves models vulnerable to membership inference attacks \cite{privacy_risks_explanations}. Additionally, adversaries can use gradient-based \cite{model_reconstruct_explanations} or counterfactual \cite{aivodji2020model} explanations to help them build highly faithful replicas of the models. 

Moreover, Shapley values have been used to identify relevant features for model agnostic backdoor poisoning attacks \cite{explanation_backdoor_attack}. One paper introduced three counterfactual explanation techniques to perform adversarial, membership inference, poisoning, and model extraction attacks on real world data sets and models \cite{xai_cybersecurity}. Alarmingly, explanations can also be used to construct attacks against ML based identity authentication protocols such as host fingerprinting and biometric-based systems \cite{garcia2018explainable}. 

\subsection{Explainability and Fairness} 

\vspace*{10px}

\textit{Post-hoc explainers should be useful for diagnosing unfairness, but often are not.}

\vspace*{10px}

The major tensions between these domains stems from the utilization of one in an effort to achieve the other. Specifically, explanations as a form of transparency and trust should be intuitive indicators for whether a system is fair \cite{Marcinkowski2020ImplicationsOA}. Adverse action notices, explanations of adverse credit scoring decisions for consumers, are justified in regulations as a method of preventing discrimination \cite{Selbst2018}. However, the reliability of explainers for this pursuit is a subject of debate with recent work \cite{begley2020explainability} outlining how they are undependable indicators of fairness. This is also affected by the difficult choice of which explainer to use \cite{Dodge2019}. Furthermore, explanations can fairwash, or promote the false perception that an ML model respects ethical values \cite{aivodji2019fairwashing}. This would essentially leave affected groups not only discriminated against, but also with no path to use explanations to contest the outcome \cite{aivodji2021characterizing}. Recent work has shown it is possible to train a model to explicitly commit fairwashing and conceal discriminatory behavior from being picked up by LIME or SHAP \cite{fooling}.

\vspace*{10px}

\textit{Post-hoc explanation methods themselves can be unfair.}

\vspace*{10px}

Further questions arise when the fairness of the produced explanations is examined. Specifically, explainability methods may exacerbate the unfairness behind algorithms by working better for certain subpopulations than others \cite{road_is_paved, dai_explanation_quality}. Further, these explanations do not necessarily preserve the fairness definitions the model is trained on \cite{dai2021will}.

\section{Context-dependent Consequences}
\vspace*{10px}
\label{sec:forward}

\subsection{What's missing? A contextual understanding.} 




Hitherto, we've described three pillars of accountability and their technical cracks: inconsistencies within themselves, mismatches with human values, and unintended consequences when they are operationalized. Furthermore, these pillars don't work well together. We've described the compounded incompatibilities that result when multiple pillars are employed. These formulas are attempts to concretize specific subjective notions of human values. They were formulated, tested, and adopted by the AI research community, a miniscule population compared to the 8 billion people on Earth \cite{PopulationClock} who may be affected by algorithmic decision-making. Ethics are fuzzy, and determining ``what is ethical" is inherently a disputable endeavor. Our mathematical formulations have deceived us into believing the morality of a technology is a measurable construct. While we may be able to achieve 100\% on a Demographic Parity score, there is no such thing as an ethics score that can be achieved at 100\%. 


In fact, in our communal endeavour of codifying proxies for human values, we have failed to properly acknowledge that there is no universally agreed upon set of moral, human values. Rather, calculating these notions through technical formulas and mathematical proofs has deceived us \cite{green2018myth} into believing that we can avoid this (and other complexities of reality) under the veil of scientific objectivity \cite{birhane2022values}.
How do we meaningfully grapple with our shortcomings without falling into the abyss of relativist debate? The first step is to acknowledge that every assumption, every decision in implementing these formulations is a choice. We must become aware of the assumptions underlying our production and claims of knowledge. These choices can only be properly evaluated when considering them within their context of deployment.




\vspace*{10px}

\textit{Ethical solutions should not be domain agnostic.}

\vspace*{10px}

For this piece, we define context as the setting in which a technology is to be deployed, and the social, political, institutional, financial, and historical influences at play in the setting. How can context-based evaluation be accomplished when, alarmingly, every discussed definition described in every pillar is domain agnostic? Every mathematical formulation does not take into account any aspects of the context under which it is being utilized. Assuming that the same fundamental property should be optimized no matter the context oversimplifies the complicated nature of reality. The inconsistencies previously described between the intended real-world outcomes and actualized real world behavior are in part due to the lack of domain consideration allotted in these formalizations. They fail to acknowledge the trade-offs, consequences, and ethical choices that are implicitly being made. Just because a property \textbf{can} be uniformly calculated in every scenario does not mean that it \textbf{should} be optimized in every scenario. In \cite{fairness_and_abstraction}, the authors describe this development of context-agnostic ethical notions as the portability trap. 

We must examine the ramifications of our choices in context. We cannot absolve ourselves of grappling with the societal impacts of the technology we build by simply implementing popular definitions of value proxies in our technical silos of academia and industry. Context is the material that maps decisions to consequences. 

\vspace*{10px}

\textit{Context is already heavily considered in other fields of ethics.}

\vspace*{10px}

The importance of context in navigating ethical decisions has precedent in more developed areas of study. Fields of bioethics, biomedical ethics, and medical ethics are built on contextual considerations. For example, the types of patient information a doctor can access are different depending on the physical context the doctor is in---if they are in a hospital versus in their car on the way to the hospital. Capturing biometric data has different ethical concerns depending on the social and institutional context of that action---is it physiological function monitoring for a patient in the ICU or the passive collection of mass amounts of physiological and behavior indicators from smartphones and digital wearables for digital phenotyping \cite{bioethics1}. In medical ethics, contextual features---professional, family, religious, financial, and institutional factors---affect what clinical decisions are made. In their training, clinicians are specifically taught to consider these as they formulate treatment plans \cite{bioethics_chapter}.

While there is much precedent for exploring tensions within pillars (Section \ref{sec:pillars}) and between pillars (Section \ref{sec:tensions}), there is a huge gap in the technical research corpus for understanding the contextual consequences that arise. We begin to address this gap by outlining key areas of consideration, highlighting three useful sociotechnical frameworks, and posing open questions for practical implementation. We push for greater contextual understandings of the impacts of technological embeddings in academia and industry.





\subsection{Examining the Real World Impact}  
\vspace*{10px}

\textit{We must recognize that our modeling assumptions do not justly reflect reality.}

\vspace*{10px}

The first set of choices to examine are the assumptions made by our modeling approaches. For instance, most fairness work considers a static world, with one population being passed through one model during one time period. This does not account for feedback loops or long-term effects. Dynamical systems offer a potential approach to understanding long term implications, by modeling the evolution and effects of fairness on a particular system over multiple time steps \cite{causal_fairness_dynamics, fairness_without_demographics, feedback_loops}. Dynamical modeling explicitly expands assumptions to more closely align with how real-world algorithmic systems might shape their environments over time.

Additionally, most fairness work either considers privilege as a binary (either you belong to the privileged group or you do not) or views it in a siloed fashion along only one demographic axis (e.g. optimizing along gender, or along race). Work in subgroup fairness helps to outline some of the limitations in this approach \cite{preventing_fairness_gerrymandering}, however, to fully embrace the intersectionalist nature of individuals, we can further question these classification systems, especially that of the male/female binary \cite{data_feminism}. This succeeds only in tandem with greater sociocultural data collection \cite{sociocultural_data} and data disaggregation \cite{kauh2021critical} to allow a broader range of demographic identities to be captured in data collection stages of the machine learning pipeline \cite{framework_bias_harms}.

\vspace*{10px}

\textit{We need to embrace viewing technology through the lens of sociotechnical systems.}

\vspace*{10px}

More broadly, to have full contextual impact awareness, we must actively consider the contexts that we deploy our technology in as relevant parts of the design process and understand the needs and wants of all of the system's stakeholders. Science and Technology Studies examines the social contexts in which technology is produced, evaluated, and deployed. The term sociotechnical system aims to describe the complex interplay between technical and human actors in real world arrangements \cite{douglas2012social}. Through the lens of sociotechnical systems, we can more meaningfully consider the ramifications and effectiveness of our technical solutions. This means asking questions like: Who are the different stakeholders in each system---the users, the practitioners, the affected communities? What does each stakeholder want from the technology? How is our technology being utilized differently by the different human stakeholders? What is the relevant historical and cultural context? Below, we outline a few useful frameworks. We acknowledge that this is not an exhaustive list, but one that sets the groundwork for the future we want to see. 

\vspace*{10px}

\textit{Three context-first reformulations already exist.}

\vspace*{10px}

The \EMPH{fairness} definitions described so far restrict analysis to isolated decisions. Instead, \cite{escaping_impossibility} proposes \EMPH{substantive algorithmic fairness}. This involves identifying structural responses for embedding fairness and analysing the hierarchies and institutional structures that surround particular decision points. Specifically, this is composed of three steps: ``1) diagnosing the substance of the inequalities in question, 2) identifying what reforms can remediate the substantive inequalities, and 3) considering whether algorithms can enhance the desired reforms."

\EMPH{Contextual Integrity} reimagines what it means to ensure \EMPH{privacy}. The theory defines privacy as the ``appropriate flow of information,'' where what is appropriate entirely depends on the context being considered. To determine what is private, one must understand who the stakeholders involved in the flow of information are, what types of information are being transmitted, and how they are being transmitted \cite{privacy_in_context}.

To address issues with \EMPH{transparency}, we must build \EMPH{domain-specific transparency methods}. In every context, we must first understand what types of transparency are useful and relevant and acknowledge how this answer varies for each stakeholder \cite{interp_stakeholders}. Then, appropriate forms of transparency should be designed according to situational needs. These methods must then be explicitly evaluated for comprehension and utility through practitioner user studies, think-aloud interviews, and feedback from relevant stakeholders \cite{judges_counterfactuals}. Only through more deeply considering the context of deployment can human-centered methods be developed \cite{explainable_ai_theory_driven}. These are already the norm in settings such as healthcare and life sciences, where methods must be developed to explain models to domain experts in very specific ways to ensure trust and adoption \cite{Kiseleva2022}.
    
While we found many resources from the computer science literature aimed at analysing technical tensions within and between pillars, to our knowledge, we were not able to identify examples of real-world rigorous contextual impact assessments. Many of these frameworks are not new (e.g. Contextual Integrity was introduced in 2010 \cite{privacy_in_context}), but there are gaps in the acceptance and employment of these strategies in practice. We push for these gaps to be reconciled and advocate for continued collaboration with sociotechnical scholars.

\subsection{Open Questions for Practical Implementation}
We've highlighted a few frameworks from academic literature, but how can we practically develop and implement contextually aware tools? To tackle this question, we identify key engineering challenges that will need to be addressed. This is not an exhaustive list of concerns, but a starting point for broader context-forward redirections.

\vspace*{10px}

\textit{How should information be collected by a contextual system? }

\vspace*{10px}

One way of viewing contextually-aware frameworks is that they ask researchers and practitioners to build systems that incorporate more information. The idea being that more information will help the system adjust to the context accordingly. However, we caution against the immediate assumption that more data is better.
Public distrust in commercial data collection is strong \cite{PEW1}, and the kinds of information that need to be collected must be justified based on specific framework requirements.

How should this information be collected and stored? In a productionalized system, we need to contend with storage requirements, standardized data formatting, and pipelines. Moreover, how we collect the data is just as important. Crafting usability studies to see which methods invoke the least friction while also designing the requirements for what a system should do to store such information will be vital.

\vspace*{10px}

\textit{What types of tools need to be developed?}

\vspace*{10px}

Through what type of format can one operationalize contextual understanding? For inspiration, we look towards tools developed in the fairness space. These tools formalize considerations in actionable ways: through checklists, thought activities, and models of understanding that are ready for immediate integration into industry workflows. Some examples include frameworks for identifying all sources of bias in machine learning pipelines \cite{framework_bias_harms}, DrivenData’s ethics checklist \cite{deon}, datasheets for datasets \cite{datasheets}, and model cards \cite{modelcards}. Moreover, we can look to Explainability Case Studies \cite{explainability_user_studies} to see how to incorporate stakeholder feedback, so that, when we design technologies, user experience aligns with user expectations.

Building frameworks and evaluation methods for contextual systems will allow us to operationalize such systems, much like we have already done with current, context-agnostic formulations. Establishing how we are building and evaluating context-aware systems could allow us to measure the long term effects these systems will have. This is extremely important for mitigating further harm.

\vspace*{10px}

\textit{How should machine learning systems respond to context?}

\vspace*{10px}

This is probably the most essential question on this list, and it can be interpreted and investigated in multiple ways. We can first read this as: what mechanism should machine learning systems use to respond to context? Should it be a team that evaluates and audits a system based on some protocol? Or maybe it should be a set of triggers that flexibly respond to context with different definitions? Another way to read this question is: how should the user experience the response of the system? This would require user studies on specific system designs and mechanisms. 

For inspiration, we can look towards \cite{nlp_unit_tests}, which offers a unit testing framework for assessing bias in natural language processing systems. This would allow for end users in collaboration with companies to generate new tests for their specific use case, that could be used in pre-production or productionalization. Moving away from static benchmark tests to curated tests for domain-specific issues is a step in the right direction.

\vspace*{10px}

\textit{What aspects of ethical responsibility does each stakeholder carry?}

\vspace*{10px}

Technology is built and deployed through complicated systems involving a variety of stakeholders: technologists, business leaders, compliance officials, etc. The types of responsibilities of each level must be identified based on situational needs. What types of contextual understanding might a model builder need to have versus a model deployer? These types of decisions might be in the realm of a new vertical within industry. Just as chief data ethics officers have been introduced \cite{chief_data_ethics}, we may need to build out a workforce that can further inform domain-specific solutions.

Using Contextual Integrity as Privacy (CI) as an example, this could look like having an employee who is responsible for collecting the parameters of CI (sender, recipient, subject, information type, and transmission principle) and creating a report to inform the types of privacy requirements necessary for specific projects. This is an overly simplistic implementation to satisfy CI, but we can imagine a world where the contextual information collection may be a necessarily manual process.

\vspace*{10px}

\textit{How can we design inclusively?}

\vspace*{10px}

During the design process of technology, inclusion needs to be prioritized. Participatory Design advocates for meaningful engagement with domain experts, end users, and any other affected communities, so that their perspectives are thoughtfully reflected throughout the development and deployment process \cite{participatory_design}. It must be ensured that this type of community involvement is not just exploitative ``participation-washing," but rather a genuine and long-term collaboration \cite{participation-washing}.

\subsection{Moving Forward}

\vspace*{10px}

\textit{Why we don't discuss accuracy tradeoffs.}

\vspace*{10px}

Throughout this paper, we have chosen to not focus on potential accuracy tradeoffs with fairness, privacy, or transparency \cite{goyal2013accuracy, pmlr-v119-dutta20a, acc_exp_tradeoff_doesnt_exist}. Debates about the fears of ``sacrificing accuracy" miss the point of embedding ethical values in our systems. It is crucial that more than just accuracy is optimized as our objective metric. By framing these notions as a zero-sum game with accuracy, data scientists are not incentivized or expected to meaningfully consider reformulations as suitable real world solutions \cite{blackboxbad}. Moreover, target labels in datasets often represent constructs, such as risk scores for recidivism, socioeconomic status, etc. These are representations that cannot be directly measured in the real world, and as a result, their representation in a dataset is fundamentally imperfect. This results in a mismatch between the theoretical understanding of the construct and how it is utilized in practice \cite{measurement_and_fairness}. Moreover, claims of accuracy are often unverifiable. It is impossible to calculate new, independent accuracy values in impactful algorithmic systems when most or all people are affected by the results. The counterfactual data on what would have happened had e.g. someone been given a loan is simply not available for measuring \cite{Grill2022}.

Further, we take aim at the community's framings of technical incompatibilities as "impossibility theorems". This choice of language normalizes researchers to view their shortcomings of accountability with ``resigned inevitability" \cite{green2018myth}.

\vspace*{10px}

\textit{Technology carries power.}

\vspace*{10px}

There is urgency in addressing our failures in properly embedding human values in machine learning systems. Technology is inherently value-laden and implicitly political \cite{moral_cryptography, data_science_political_action}. The use of technological solutions redistributes power---who gets to make decisions and what information is made accessible for those decisions.

With stakes this high, we must recognize that technology is not always the solution. Substantive algorithmic fairness argues that a key step in a structural ethical response is to critically consider whether algorithms can enhance or facilitate the necessary reforms \cite{escaping_impossibility}. A failure to recognize the possibility that the best solution to a problem may not involve technology leads to the so-called solutionism trap \cite{fairness_and_abstraction}. Yet, technology can still be extremely valuable in specifically-scoped, context-aware roles, such as a tool for measuring social problems, for defining social problems, for clarifying the limits on technical interventions, and for highlighting social problems in novel ways \cite{social_change_computing}.

\section{Conclusion}
\vspace*{10px}
The current formalisms adopted by the AI community for embedding ethical values are severely lacking. Popular notions of fairness, privacy, and model transparency each carry their own inherent tensions, as well as additional tensions when multiple pillars are employed in tandem. These pillars also suffer from a portability trap and a lack of awareness for the context in which the technology is being implemented \cite{fairness_and_abstraction}. Because of this, they fail to acknowledge the trade-offs, consequences, and ethical choices that are implicitly being made.  Context is the material that maps decisions to consequences. We cannot continue to use these mathematical formalizations to avoid grappling with the real-world impacts of technology. We push for greater emphasis on implementing contextually aware technical interventions for accountability.

\section*{Acknowledgments}
%

We greatly appreciate the advice and support from Lizzie Kumar, Avi Schwarzschild, and Yaniv Yacoby. In particular, we would like to thank Valentine d'Hauteville and Sarah Ostermeier for their extensive feedback on our work.








\bibliographystyle{IEEEtran}
\bibliography{main}

\newpage

\appendix
\subsection{Technical Definitions of Fairness}
\begin{table}
    \centering
    \caption{Five popular fairness definitions: $Y$ represents a binary ground truth label, $\hat{Y}$ represents a binary prediction, $A$ represents a protected attributes, and $R$ represents a score. The criterion type captures the class of fairness definition.}
    \begin{tabular}{|c|c|c|}
        \hline
        Name & Probabilistic Definition & Criterion Type \\
        \hline \hline
         Demographic Parity \cite{dem_par} & $P(\hat{Y} = 1 | A = a) = P(\hat{Y} = 1 | A = b)$ & Independence \\
         \hline
         80\% Rule \cite{80_rule} & $\frac{P(\hat{Y} = 1 | A = a)}{P(\hat{Y} = 1 | A = b)} \geq 0.8$ & Independence \\
         \hline
         Equal Opportunity \cite{eq_odds} & $P(\hat{Y} = 1 | Y = 1, A = a) = P(\hat{Y} = 1 | Y = 1, A = b)$ & Separation \\
         \hline
         Equalized Odds \cite{eq_odds} & \shortstack{$P(\hat{Y} = 1 | Y = 1, A = a) = P(\hat{Y} = 1 | Y = 1, A = b)$ \\ $P(\hat{Y} = 1 | Y = 0, A = a) = P(\hat{Y} = 1 | Y = 0, A = b)$} & Separation \\
         \hline
         Calibration of Groups \cite{cal} & $P(Y = 1 | R = r, A = a) = r $ & Sufficiency \\
         \hline
    \end{tabular}
    \label{tab:fair_defs}
\end{table}

Table~\ref{tab:fair_defs} outlines five popular fairness definitions: $Y$ represents a binary ground truth label, $\hat{Y}$ represents a binary prediction, $A$ represents a protected attributes, and $R$ represents a score. The criterion type captures the class of fairness definition.  We redirect the reader to~\cite{Verma2018} and~\cite{fairml} for additional discussion about and formalization of fairness definitions.

\subsection{Technical Definitions of Differential Privacy}
\label{sec: defprivacy}
\begin{definition}[$\epsilon$-Differential Privacy \cite{diff_priv_sem}]
For any $\epsilon > 0$, a randomized algorithm $f$ satisfies $\epsilon$-Differential Privacy if for any pair of neighboring datasets $D$, $D'$ and for all $S \subset$ Range($f$)

$$P(f(D) \in S) \leq e^{\epsilon} P(f(D') \in S)$$
\end{definition}

A relaxation of this definition was created soon after, which loosens the probabilistic restriction of the $e^{\epsilon}$.

\begin{definition}[$(\epsilon, \delta)$-Differential Privacy \cite{diff_ed}]
 For any $\epsilon, \delta > 0$, a randomized algorithm $f$ satisfies $(\epsilon, \delta)$-Differential Privacy if for any pair of neighboring datasets $D$, $D'$ and for all $S \subset$ Range($f$)

$$P(f(D) \in S) \leq e^{\epsilon} P(f(D') \in S) + \delta$$ 
\end{definition}

\end{document}